\documentclass[times, review, 10pt]{elsarticle}

\usepackage{caption}
\usepackage{amssymb}

\usepackage[font=small,labelfont=bf,labelsep=none]{caption}
\usepackage{amsmath,amsfonts}
\usepackage{algorithmic}
\usepackage{algorithm}
\usepackage{array}
\usepackage{textcomp}
\usepackage{stfloats}
\usepackage{url}
\usepackage{verbatim}
\usepackage{graphicx}
\usepackage{booktabs}

\usepackage[caption=false,font=scriptsize,labelfont=rm,textfont=rm]{subfig}
\usepackage{color} 
\usepackage{graphicx}
\usepackage{multirow}
\usepackage{booktabs}
\usepackage{bm} 

\journal{Pattern Recognition}

\begin{document}

\begin{frontmatter}

\title{iFADIT: Invertible Face Anonymization via Disentangled Identity Transform}



\author[aff1]{Lin~Yuan}
\author[aff1]{Kai~Liang}
\author[aff1]{Xiong~Li}
\author[aff1]{Tao~Wu}
\author[aff2]{Nannan~Wang}
\author[aff1]{Xinbo~Gao}
\affiliation[aff1]{organization={Chongqing Key Laboratory of Image Cognition},
            addressline={Chongqing University of Posts and Telecommunications}, 
            city={Chongqing},
            postcode={400065}, 
            state={Chongqing},
            country={China}}
\affiliation[aff2]{organization={State Key Laboratory of Integrated Services Networks},
            addressline={Xidian University}, 
            city={Xi'an},
            postcode={710071}, 
            state={Shaanxi},
            country={China}}            

\begin{abstract}
Face anonymization aims to conceal the visual identity of a face to safeguard the individual's privacy. 
Traditional methods like blurring and pixelation can largely remove identifying features, 
but these techniques significantly degrade image quality and are vulnerable to deep reconstruction attacks. 
Generative models have emerged as a promising solution for anonymizing faces while preserving a natural appearance. 
However, many still face limitations in visual quality and often overlook the potential to recover the original face 
from the anonymized version, which can be valuable in specific contexts such as image forensics. 
This paper proposes a novel framework named iFADIT, 
an acronym for Invertible Face Anonymization via Disentangled Identity Transform.
The framework features a disentanglement architecture coupled with a secure flow-based model: 
the former decouples identity information from non-identifying attributes, 
while the latter transforms the decoupled identity into an anonymized version 
in an invertible manner controlled by a secret key. 
The anonymized face can then be reconstructed based on a pre-trained StyleGAN 
that ensures high image quality and realistic facial details. 
Recovery of the original face (aka de-anonymization) is possible upon the availability of the 
matching secret, by inverting the anonymization process based on the same set of model parameters. 
Furthermore, a dedicated secret-key mechanism along with a dual-phase training strategy 
is devised to ensure the desired properties of face anonymization. 
Qualitative and quantitative experiments demonstrate the superiority 
of the proposed approach in anonymity, reversibility, security, diversity, 
and interpretability over competing methods. 
\end{abstract}

\begin{keyword}
Privacy protection, face anonymization, disentanglement, flow-based model.
\end{keyword}

\end{frontmatter}

\section{Introduction} \label{sec:introduction}

Face capturing has become a ubiquitous activity across a wide range of applications, 
including face recognition, video surveillance, and healthcare. 
However, the extensive collection of facial data poses a potential threat to individual privacy. 
Unlike traditional revocable digital secrets, such as passwords, facial data is one of the most 
sensitive biometric identifiers due to its uniqueness and immutability. 
Once stolen or compromised, ``resetting'' one's facial data is virtually impossible, 
leading to potentially severe consequences from malicious use. 
Consequently, safeguarding facial privacy has become an extremely crucial yet challenging task.

\begin{figure}[ht]
\centering
\includegraphics[width=\columnwidth]{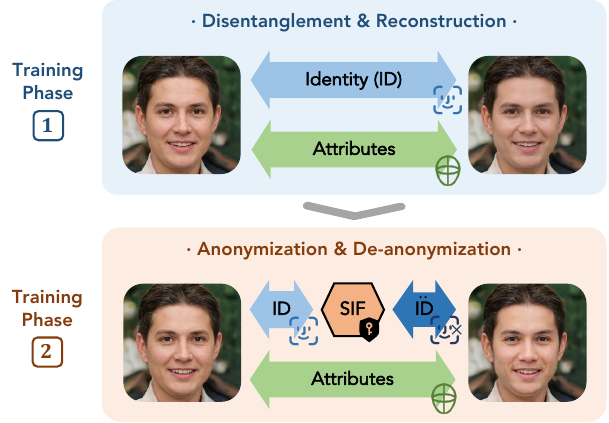}
\caption{The paradigm iFADIT built on a dual-phase training strategy: 
The first training phase optimizes a disentangle-reconstruct architecture 
that tries to decouple the identity and non-identifying representations of a face image. 
The second phase focuses on training a secure identity flow-based (SIF) model 
capable of transforming the disentangled identity representation 
in an invertible way controlled by a user-specific secret. 
}
\label{fig:paradigm_v2}
\end{figure}

A considerable number of face anonymization techniques have been developed to 
protect facial privacy in various scenarios. 
Early obfuscation-based methods
~\cite{yuan2017iet,erdelyi2014cartoon,zhou2021pixelate,chattopadhyay2007cvpr,
zhang2022tpe,dufaux2008scramble,agrawal2011tcsvt,sohn2011tcsvt,ciftci2018falsecolor}, 
such as blurring, pixelation, and scrambling, can efficiently obscure visual information of a face, 
but result in images with low visual utility and camouflage capabilities, 
and are prone to reconstruction attacks.  
With the advancement of deep generative models, generation-based face anonymization 
has emerged as a promising alternative for protecting facial privacy
~\cite{maximov2020ciagan,hukkelas2019deepprivacy,
gu2020eccv,proenca2022uunet,yuan2022ivf,yuan2022proface,wen2022idmask,zhang2023rapp,li2023riddle}. 
These methods attempt to morph the original face into a different one, 
resulting in a natural facial appearance that presents an anonymized identity.  
However, two common problems still exist for this branch of methods: 
1)~Anonymized faces generated by most existing methods show limited image quality, 
especially with visible manipulation artifacts or domain disparity from the original. 
2)~Most methods focus on the process of anonymization but overlook the de-anonymization capability, 
resulting in limited de-anonymization fidelity, security, or model interpretability.

In response to above challenges, we attempt to explore a novel framework for face anonymization 
with a particular emphasis on enhancing the generated image quality and ensuring secure reversibility. 
The framework is named {\bf iFADIT}, an acronym for {\bf i}nvertible {\bf F}ace 
{\bf A}nonymization via {\bf D}isentangled {\bf I}dentity {\bf T}ransform.
Central to iFADIT is a disentanglement architecture 
that decouples the identity and non-identifying information of a face image, 
coupled with a pre-trained generation model (i.e., StyleGAN~\cite{karras2019style-ffhq}) for reconstructing 
high-quality face images from the disentangled representations. 
The core objective of invertible face anonymization is achieved by transforming 
the disentangled identity representation in a secure and invertible manner, 
powered by a specialized flow-based model integrated with secret keys. 
To ensure the expected properties of anonymization and de-anonymization, 
a dual-phase training strategy is proposed to optimize the disentanglement 
and transformation processes progressively. 
An abstract illustration of this framework is shown in Fig.~\ref{fig:paradigm_v2}. 
The contributions of the paper can be summarized as follows:
\begin{itemize}
    \item A novel framework for face anonymization is proposed, 
    which achieves secure and invertible anonymization 
    by transforming the facial identity representation 
    disentangled in the latent space of StyleGAN. 
    The disentanglement architecture involved enables a clean isolation 
    of sensitive identity information from non-identifying attributes, 
    such that the facial attributes (e.g., posture and expression) keep intact 
    during face anonymization. 
    The StyleGAN applied ensures high-quality face image reconstruction, 
    taking advantage of the knowledge learned about human face from massive image data. 
    \item  Building upon the disentangle-reconstruct architecture, 
    a secure identity transform module is devised based on a flow-based model, 
    with the aim of transforming the identity representation into anonymized forms 
    according to user-specific secrets, in an invertible and more interpretable manner.
    Beyond, a dual-phase training strategy along with comprehensive loss functions 
    is proposed to maximize the expected functioning of the proposed approach. 
    \item Qualitative and quantitative evaluations conducted on multiple 
    face image datasets are provided, demonstrating the superiority of the 
    proposed approach over competing methods in various aspects including 
    identity anonymity, image quality, and reversibility. 
\end{itemize}


\section{Related Work} \label{sec:related_work}

\subsection{Face Anonymization}

Generative face anonymization has emerged as a more promising solution, 
which morphs a given face into a different one, 
maintaining a natural facial appearance while exhibiting an 
anonymized identity distinct from the original.
DeepPrivacy~\cite{hukkelas2019deepprivacy} employs a conditional generative adversarial network 
that utilizes facial key points as the condition to guide the anonymized faces 
preserving the original pose and image background. 
Similarly, CIAGAN~\cite{maximov2020ciagan} utilizes an one-hot vector signaling the 
condition of facial identity to generate diverse anonymized faces. 
Kuang et al.~\cite{kuang2021acmmm} proposed DeIdGAN, a de-identification GAN-based model 
for face anonymization by replacing a given face image with a different synthesized yet realistic one. 
Gafni et al.~\cite{gafni2019iccv} first considered face anonymization for sequential video frames, 
using a feed-forward encoder-decoder network conditioned on the high-level facial representation. 
Yet, the above methods did not take reversibility into account, 
and the resulting image quality is relatively limited.  

Reversible approaches to face anonymization have been studied in response to 
use cases where original faces are needed for image analysis or forensics. 
Gu et al.~\cite{gu2020eccv} proposed a reversible face identity transformer, aka FIT,  
capable of both anonymization and de-anonymization. In FIT, a secret key is integrated as 
the condition of a generation model such that different keys result in distinct anonymized 
identities, and that one can recover the original face only with the matching secret key. 
UU-Net~\cite{proenca2022uunet} presents a reversible framework for facial privacy protection 
in the context of video surveillance, where two U-Nets~\cite{Olaf2015UNet} 
are utilized for protection and restoration respectively. 
Recently, Li et al.~\cite{li2023riddle} proposed RiDDLE, a reversible de-identification framework 
implemented by ``encrypting'' the latent input of StyleGAN~\cite{karras2019style-ffhq}. 
Wen et al.~\cite{wen2022idmask} proposed IdentityMask, a modular architecture for reversible 
face video anonymization that leverages deep motion flow to avoid per-frame evaluation. 
However, most of those approaches show limited de-anonymization quality, namely, 
the similarity between the recovered and original images in terms of both 
human visual perception and identity feature space still has a room to improve. 

Several methods attempt to achieve a trade-off between utility and privacy 
of face anonymizaiton so that anonymized images can still fulfill certain 
machine vision tasks, such as identity recognition. 
Li et al.~\cite{li2021acmmm,li2023tifs} proposed identity-preserved face anonymization, 
which generates visually anonymized faces that can still be recognized by machines, 
relying on an identity-aware region discovery module to determine the 
facial regions sensitive to humans but less crucial to machines.  
IVFG~\cite{yuan2022ivf} generates exclusive virtual faces for each user,
based on which, identity recognition can be accurately performed 
relying on those virtual faces. 
Yuan et al.~\cite{yuan2022proface} proposed PRO-Face, which tries to ``embed'' 
the original face into its pre-obfuscated form in such a way that a face recognition 
model can ``extract'' the original identity hidden in the obfuscated face. 
Zhang et al.~\cite{zhang2023rapp} proposed a reversible solution for facial attribute anonymization, which supports recognizing the real identity from the anonymized face.

Frankly, we are not the first applying disentanglement architecture in face anonymization. 
Cao et al.~\cite{cao2021iccv} proposed an invertible anonymization framework 
where anonymization and de-anonymization are achieved by transforming the 
disentangled identity via vector composition, according to a user-specified password. 
A multi-level attribute encoder and a pre-trained identity encoder are utilized 
to obtain the disentangled identity and attribute representations. 
AdaDeId~\cite{ma2022adadeid} presents a method that can continuously manipulate 
the disentangled identity based on a unit spherical space, to achieve various 
de-identified faces.
Recently, StyleID~\cite{le2023styleid} provides an anonymization framework that 
transforms facial identity disentangled in the latent space of StyleGAN, 
while preserving as many characteristics of the original faces as possible. 
Yet, the method relies on a target face image for identity transformation, 
which may affect the privacy of another person. 
Besides, it does not support de-anonymization. 
Although disentanglement-based approaches provide a fresh and more interpretable view 
for face anonymization, most methods in this branch still lack a sophisticated mechanism 
to disentangle and transform the identity information effectively, 
resulting in limited anonymization or de-anonymization performance.



\subsection{Flow-based Model}
Flow-based models, also known as invertible neural networks (INN), is a type of generative model  
that learns to transform a simple probability distribution, such as a Gaussian, 
into a more complex distribution that matches the data being modeled. 
This transformation is achieved through a series of bijective functions, 
which are intrinsically invertible, allowing for both efficient sampling 
and exact computation of the likelihood of the data. 
Their invertibility also ensures that they can map complex data distributions back to simpler ones, 
facilitating tasks like data generation, density estimation, and data manipulation. 
Examples of early-stage flow-based models include Non-linear Independent Components Estimation~(NICE)~\cite{dinh2014nice} 
and RealNVP~\cite{Realnvp}, followed by Glow~\cite{kingma2018glow} that supports invertible 1$\times$1 convolutions 
for more efficient and realistic image generation. 

The inherent invertibility of the flow-based model motivates us to employ it for 
transforming facial identity information, allowing for easy recovery of the original identity.
Thanks to the bijective structure utilized, the invertible transformation of identity 
information in a flow-based model is considered more interpretable~\cite{Realnvp}. 
To ensure the security of invertible identity transformation, 
we adopt a conditional invertible neural network (cINN) as proposed by~\cite{ardizzone2021cinn}. 
In this model, a conditional signal is concatenated into the feature maps within each bijective 
function to guide the generation of images with diverse styles. 
Similarly, we integrate a secret key as a condition into the flow-based model 
to ensure not only diverse anonymizations but also secure de-anonymization 
that can be achieved only with the matching secret key.


\section{The Proposed Approach} \label{sec:method}
\subsection{Overview}
\paragraph{Thread model and privacy protection target}
We consider a privacy protection scenario where face anonymization is applied to 
\textit{prevent the identity information of the original face from being recognized by both machines and humans 
while preserving non-identifying facial attributes (such as outline and pose) as much as possible}. 
Leveraging the success of recent face recognition models in replicating human vision's capability 
to distinguish facial identities, we utilize pre-trained face recognition models to represent facial 
identity information and to evaluate the anonymization performance throughout the study. 
To meet the needs of security and utility, we aim to control the anonymization process using a secret key. 
This approach ensures that different keys produce different anonymization results, similar to data encryption. 
In cases of forensics or analysis, the original face can be securely recovered from the anonymized one 
by using the matching secret key, thereby simulating a decryption process.

\begin{figure}[t]
 \centering
 \includegraphics[width=\columnwidth]{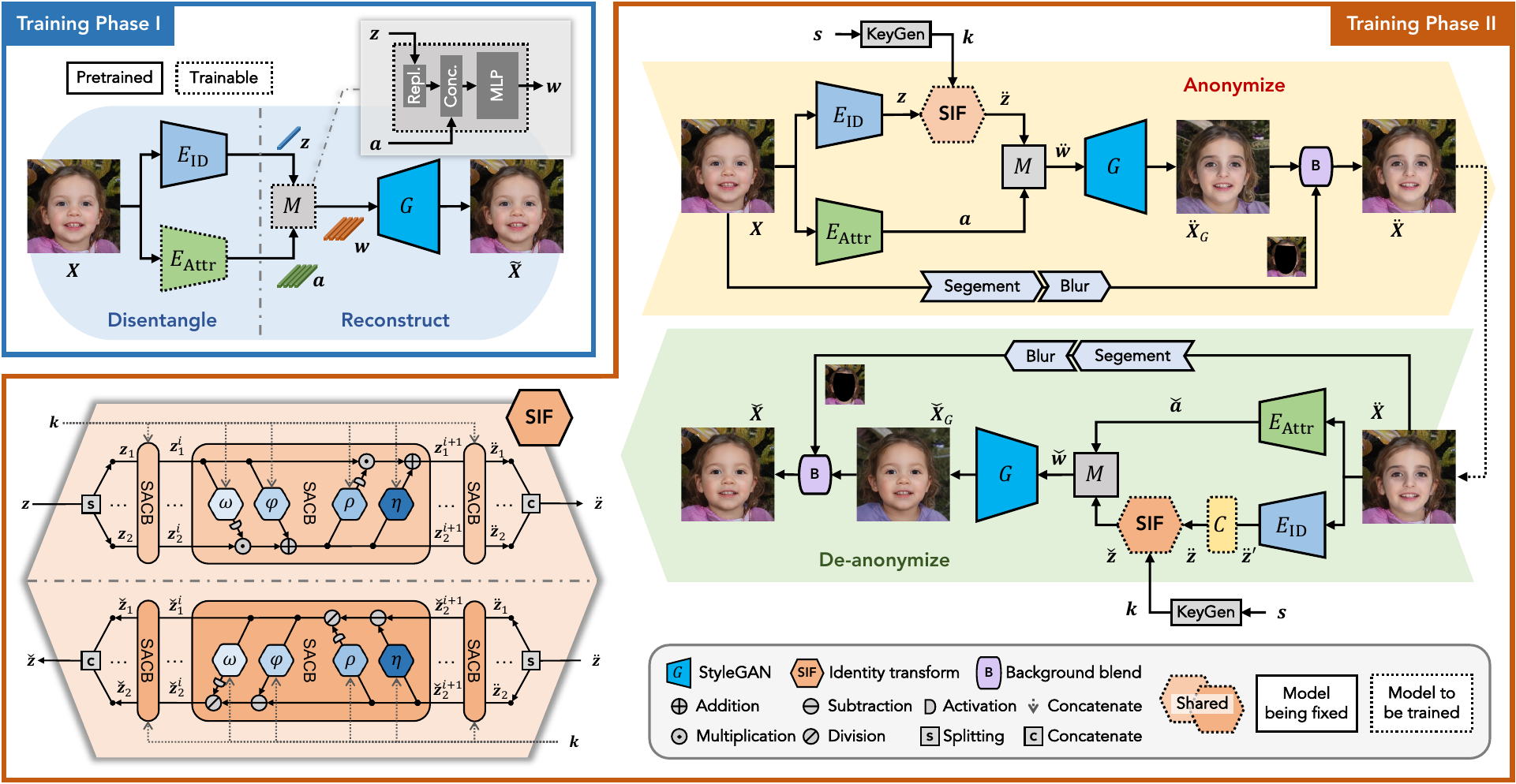}
 \caption{The overall framework of iFADIT, which consists of two training phases. In the first training phase, the identity disentanglement and image reconstruction architecture is built. In the second training phase, the introduced SIF and ICL are imported and optimized, with other components fixed. }
 \label{fig:framework}
\end{figure}

\paragraph{Core designing concept}
The above-mentioned objective of face anonymization motivates us 
to decouple the identity and attribute information from a face image, 
and then alter the identity solely to differentiate it from the original. 
To this end, a disentanglement-reconstruction architecture is first devised to disentangle 
the identity representation and non-identifying attribute representation from a face image, 
based on which, the original face image can be reconstructed with high fidelity utilizing 
a pre-trained face generation model. 
Next, face anonymization involves transforming 
the disentangled identity representation into a new one while leaving the non-identifying attributes unchanged. 
By executing the reconstruction module on the transformed identity and the original attributes, 
the reconstructed image presents a natural face with an anonymized identity different from the original, 
maintaining a similar facial pose and outline.
The identity transformation is performed by a flow-based model with intrinsic invertibility, 
controlled by a user-specific secret key. Using the matching secret key, the original identity 
can be recovered from the anonymized face through the same transformation model. 
Since the non-identifying attributes are supposed to be unchanged during face anonymization, 
the same attributes representation can be obtained from the anonymized image. 
By feeding the recovered identity and attribute representations back into the reconstruction module, 
one is able to obtain the de-anonymized face that closely mirrors the original. 
A specialized secret-key mechanism is devised such that different keys produce diverse anonymization results, 
and only with the matching secret key is it possible to recover the de-anonymized face. 
A dedicated dual-phase training strategy is proposed to progressively optimize 
the disentanglement and transformation processes, proving to be highly effective.
The entire iFADIT framework is illustrated in Fig.~\ref{fig:framework} and 
the details of its distinct components are provided as follows. 

\subsection{Disentanglement and Reconstruction}
First, the disentanglement and reconstruction modules can be formulated as follows: 
Given a face image denoted as $\bm{X}$, we devise a disentanglement module 
$\mathcal{D}$ capable of decoupling the representation related to facial identity 
(denoted as $\bm{z}$) and non-identifying attributes (denoted as $\bm{a}$) separately, 
based on which, the original image can be reproduced by a reconstruction module denoted as $\mathcal{R}$: 
\begin{equation}
    (\bm{z}, \bm{a}) = \mathcal{D}\left(\bm{X}\right) \Longrightarrow \bm{\Tilde{X}} = \mathcal{R}\left(\bm{z}, \bm{a}\right),
\end{equation}
where $\bm{\Tilde{X}}$ denotes the reconstructed image closely mirroring the original. 

Inspired by GAN inversion techniques~\cite{psp,nitzan2020tog,xia2023ganinv}, we opt for disentangling 
the identity information in the latent space of StyleGAN~\cite{karras2019style-ffhq}, 
a representative and powerful generation model capable of synthesizing high-quality realistic faces. 
As the upper-left part (labeled Training Phase I) of Fig.~\ref{fig:framework} illustrates, 
the disentanglement module consists of a pair of encoders for extracting 
the identity and attributes representations respectively,  
denoted as $E_\mathrm{ID}$ and $E_\mathrm{Attr}$. 
Following a similar work on face identity disentanglement~\cite{nitzan2020tog}, 
we utilize a pre-trained ArcFace model~\cite{deng2019arface} as the identity encoder 
(i.e., $\bm{z} \in \mathbb{R}^{512}$). 
Instead of mapping the facial attributes into the $\mathcal{W} \sim \mathbb{R}^{512}$ 
latent space of StyleGAN (as~\cite{nitzan2020tog} does), 
we opt for using pSp~\cite{psp} for encoder $E_\mathrm{Attr}$ to obtain the 
attributes representation in the $\mathcal{W}^+ \sim \mathbb{R}^{14\times 512}$ space, 
with the aim of preserving more fine-grained facial details. 

To reconstruct the original image from the disentangled representations, 
a similar mapping network (denoted as $M$) as~\cite{nitzan2020tog} is utilized. 
It first replicates the identity vector to match the dimension of the attributes representation. 
Then, the concatenation of the two representations is fed into a two-layer MLP, 
generating the latent code $\bm{w} \in \mathcal{W}^+$: 
\begin{equation} \label{eq:m_module}
    \bm{w} = M\left( \bm{z}, \bm{a} \right) := \mathrm{MLP} \left(\left[R(\bm{z}), \bm{a}\right] \right),
\end{equation}
where $R(\cdot)$ denotes the vector replication and 
$\left[\cdot,\cdot\right]$ indicates concatenation along the vector dimension. 
Finally, applying the StyleGAN model (denoted as $G$) on $\bm{w}$, the reconstruction image 
(denoted as $\bm{\Tilde{X}}$) highly mirroring the original one  can be generated.
Since the pre-trained identity encoder and StyleGAN generator have acquired 
extensive knowledge about human faces from vast amounts of training data, 
we utilize their model parameters without updating them. 
Instead, we train the attribute encoder and the mapping network.

\subsection{Secure Identity Flow} 
Building on a pair of disentanglement and reconstruction modules, 
face anonymization can be effectively achieved by incorporating 
an identity transformation module in between, 
forming a Disentangle-Transform-Reconstruct pipeline.
The transformation module manipulates the decoupled identity representation 
using a user-specified secret key in an invertible manner. 
Specifically, different keys guide the module to produce distinct identities, 
and the original identity can be recovered using the same set of transformation parameters, 
provided the matching key is used.
This transformation is applied solely to the identity representation, 
leaving non-identifying facial attributes unaffected. 
To this end, we propose Secure Identity Flow (SIF), 
a flow-based model with secret key integrated as the condition 
to control the secure and invertible transformation of the identity representation. 
The forward ($T$) and backward ($T^{-1}$) function of SIF can be formulated as 
\begin{equation}
    \ddot{\bm{z}} = T(\bm{z}, s)  \Longrightarrow \Check{\bm{z}} = T^{-1}(\ddot{\bm{z}}, s),
\end{equation}
where $\ddot{\bm{z}}$ stands for the anonymized identity representation, 
$s$ for a user-specified secret key with arbitrary length, 
and $\Check{\bm{z}}$ for the correctly recovered identity by the inverse process of SIF.
The details of the key components of SIF are provided below:

\subsubsection{Secret Key Generation} 
First, we design a key generation module (KeyGen for short) responsible 
for generating random information to control the identity transformation. 
The KeyGen module starts with a key derivation function (KDF) that converts  
the user-specific secret $s$ with arbitrary size into a fixed-size pseudorandom byte sequence, 
followed by a $\mathrm{Norm}$ operator that linearly normalizes each byte into a float lying in the range of $[-1,1]$. 
Furthermore, utilizing the pretrained style code mapping network in StyleGAN (known as $f$), 
which maps any input noise variable into a latent style code $\bm{w} \in \mathcal{W}$, 
we obtain the final secret vector $\bm{k}$ to be used in SIF. 
Therefore, the KeyGen module can be formulated as 
\begin{equation}
\label{eq:key_gen}
    \bm{k} = \mathrm{KeyGen}\left(s\right) 
    := f \left( \mathrm{Norm}\left( \mathrm{KDF}\left( s \right)\right)\right)  \in \mathbb{R}^{512}.
\end{equation}
The KeyGen module plays two roles: 
1) It produces pseudorandom information from the user-specific secret, ensuring that any difference 
(even a single bit) in the input secret would result in completely uncorrelated pseudorandom information. 
2) It maps the pseudorandom information into a latent style code for StyleGAN, 
making the secret information adaptive to StyleGAN's reconstruction capability.

\subsubsection{Secure Affine Coupling Blocks} 
Having the secret vector $\bm{k}$ prepared, 
the SIF begins with splitting an input identity representation into two halves 
$\left(\bm{z}_1,  \bm{z}_2\right) = \mathrm{Split}\left( \bm{z} \right)$, 
which are then fed into a sequence of $N$ invertible affine coupling blocks (ACBs). 
Herein, we utilize the ACB structure proposed by RealNVP~\cite{Realnvp}, 
which is composed of four nonlinear mapping functions each structured by an MLP. 
The secret vector $\bm{k}$ derived from the KeyGen module is integrated into each ACB, 
by concatenating into the input feature of each nonlinear function in ACB, 
thus forming a so-called secure affine coupling block (SACB). 
The function of each SACB in the forward transformation pass can be formulated as:
\begin{align}
\label{eq:forward1}
	\bm{z}_2^{i+1} &= \bm{z}_2^{i} \cdot \exp \left(\mathrm{a} \left(\omega\left(\left[\bm{z}_1^{i}, \bm{k}\right]\right)\right)\right) + \varphi\left(\left[\bm{z}_1^{i}, \bm{k}\right]\right), \\
\label{eq:forward2}
	\bm{z}_1^{i+1} &= \bm{z}_1^{i} \cdot \exp \left(\mathrm{a} \left(\rho\left(\left[\bm{z}_2^{i+1}, \bm{k}\right]\right)\right)\right) + \eta\left(\left[\bm{z}_2^{i+1}, \bm{k}\right]\right), 
\end{align}
where $i \in [1..N]$ and $(\bm{z}_1^{1}, \bm{z}_2^{1})$ 
are equivalent to the splits of the original identity $(\bm{z}_1, \bm{z}_2)$. 
Notably, $\omega$, $\varphi$, $\rho$, and $\eta$ denote the above-mentioned 
nonlinear functions in each ACB, all sharing the same MLP structure. 
Each function maps the identity-secret concatenation with dimension 
$\mathbb{R}^{768}$ to a new representation with the same 
length as the input identity split in $\mathbb{R}^{256}$.
The $\mathrm{a}(\cdot)$ in Eq.~(\ref{eq:forward1}) and~(\ref{eq:forward2}) 
stands for Sigmoid multiplied by a constant factor, serving as 
a sort of activation function to constrain the processed embedding. 
The final dual output vectors of the $N$ forward SACBs are concatenated to form the output 
anonymized identity representation $\ddot{\bm{z}} = \left[\bm{z}_1^{N+1}, \bm{z}_2^{N+1}\right]$.

Due to the intrinsic invertibility of the affine coupling blocks, 
the original identity representation can be recovered from the anonymized one 
relying on the same SIF model parameters and the matching secret vector: 
\begin{align}
  \label{eq:backward1}
	\check{\bm{z}}_1^{i} &= \left(\check{\bm{z}}_1^{i+1} - \eta \left(\left[\check{\bm{z}}_2^{i+1}, \bm{k}\right]\right)\right) 
	\cdot \exp \left(-a \left(\rho \left(\left[\check{\bm{z}}_2^{i+1} , \bm{k}\right]\right)\right)\right), \\
  \label{eq:backward2}
	\check{\bm{z}}_2^{i} &= \left(\check{\bm{z}}_2^{i+1} - \varphi\left(\left[\check{\bm{z}}_1^{i}, \bm{k}\right]\right)\right) 
	\cdot \exp\left(-a\left(\omega\left(\left[\check{\bm{z}}_1^{i}, \bm{k}\right]\right)\right)\right),
\end{align}
where $\check{\bm{z}}_j^{N+1}, j \in \{1,2\}$ are equivalent to 
$\bm{z}_j^{N+1}$, the splits of the anonymized identity representation $\ddot{\bm{z}}$.
With the backward SACBs, the final dual output vectors are 
concatenated to form the recovered identity representation 
$\check{\bm{z}} = \left[ \check{\bm{z}}_1^{1}, \check{\bm{z}}_2^{1} \right]$. 

\subsection{Anonymization and De-anonymization}
To accomplish face anonymization and de-anonymization, 
the anonymized identity $\ddot{\bm{z}}$ generated by SIF along with the original attributes 
$\bm{a}$ are re-coupled and mapped into a new latent code $\ddot{\bm{w}}$ 
via a feature mapping network $M$. 
Finally, the pre-trained StyleGAN generator $G$ is applied to generate 
the anonymized face image.
The entire anonymization process can be formulated as 
\begin{equation} \label{eq:anonymization}
    \ddot{\bm{X}} = G\left( M\left( T\left(E_\mathrm{ID}\left( \bm{X} \right), s\right), E_\mathrm{Attr}\left( \bm{X} \right) \right) \right).
\end{equation}

The anonymized face $\ddot{\bm{X}}$ is supposed to present a novel identity represented 
by $\ddot{\bm{z}}$, while maintaining the non-identifying attributes of the original face, 
represented by $\bm{a}$. 
Therefore, $\ddot{\bm{z}}$ and $\bm{a}$ are expected to be extracted from the anonymized face 
via the same disentanglement module. Then, given the matching secret $s$ provided, 
the original image can be recovered by reverting the SIF module, 
re-coupling the extracted identity and attributes, 
followed by executing the StyleGAN generator again: 
\begin{equation} \label{eq:deanonymization}
    \Check{\bm{X}} = G\left( M\left( T^{-1}\left(C\left(E_\mathrm{ID}\left( \ddot{\bm{X}} \right)\right), s\right), E_\mathrm{Attr}\left( \ddot{\bm{X}} \right) \right) \right).
\end{equation}

\paragraph{Identity Compensation Layer}
The identity representation extracted from the anonymized face 
may slightly differ from $\bm{\ddot{z}}$, which may affect the recovery performance of SIF.
To mitigate this impact, we propose a learnable \textbf{\em Identity Compensation Layer (ICL)} 
(denoted as $C$), placed right after the identity encoder in 
the de-anonymization process. The ICL utilizes a four-layer MLP to correct the extracted 
identity from the anonymized face towards the one before image reconstruction, 
such that SIF can recover the original identity as expected. 

\paragraph{Facial background refinement}
As most StyleGAN inverted images can hardly maintain 
the original facial background~\cite{song2022editing-deghosting-net}, 
we introduce a refinement module that substitutes the background 
of the generated image with that of the original, using a soft bordered facial mask. 
Specifically, we first extract a binary facial mask from the input image, 
using a pre-trained BiSeNet~\cite{Yu_2018_ECCV} segmentation model. 
The mask is then Gaussian blurred, resulting in a soft mask with a gradually varied boundary, 
denoted as $\bm{M}$. 
The final anonymized image is then created by alpha blending the image just generated from StyleGAN 
(denoted as $\ddot{\bm{X}}_\mathrm{G}$) with the input image based on the mask $\bm{M}$: 
\begin{equation}
    \ddot{\bm{X}} = \bm{M} \odot \ddot{\bm{X}}_\mathrm{G} + (1 - \bm{M}) \odot \bm{X},
\end{equation}
where $\odot$ denotes element-wise multiplication propagated to each image channel. 
The same background refinement module is applied to the de-anonymization process as well. 
Note, such a module is only executed in the inference stage, without participation in training.

\subsection{Training Strategy}
To fully maximize the desired functionality, 
we propose a \textbf{\em dual-phase training (DPT)} strategy to optimize
the disentanglement and (de-)anonymization processes progressively. 
The first phase focuses on optimizing the disentanglement-reconstruction capability only, 
without the involvement of identity transformation, namely SIF. 
Then, the second phase includes SIF and ICL into training, 
while freezing the parameters of all the other components. 

\subsubsection{Training Phase I}
In the first training phase, only the attribute encoder $E_\mathrm{Attr}$ and 
the mapping network $M$ are learnable, while the other components remain fixed. 
We first define visual quality (VQ) loss to ensure the reconstructed image 
closely resembles the original image. This is achieved by using the 
Structural Similarity Index (SSIM)~\cite{wang2004ssim} and the L1 distance:
\begin{equation}
    \mathcal{L}_\mathrm{VQ} = \mathbb{E}_{(\bm{X})}\Big[\alpha \cdot (1-\mathrm{SSIM}(\bm{\Tilde{X}},\bm{X})) + \beta \cdot \| \bm{\Tilde{X}}-\bm{X}\|_1 \Big],
\end{equation}
where $\alpha$ and $\beta$ are set to 0.84 and 0.16 respectively following~\cite{nitzan2020tog}. 
We further constrain the semantic information of the reconstructed image, 
utilizing the pre-trained identity encoder~$E_\mathrm{ID}$ and a landmark extractor~$E_\mathrm{LM}$ proposed by~\cite{feng2018wingloss}:
\begin{equation} \label{eq:loss_sem}
    \mathcal{L}_\mathrm{Sem} = \mathbb{E}_{(\bm{X})} \left[ \| E_\mathrm{ID}(\bm{\Tilde{X}})-E_\mathrm{ID}(\bm{X})\|_1 +\| E_\mathrm{LM}(\bm{\Tilde{X}})-E_\mathrm{LM}(\bm{X})\|_1 \right].
\end{equation}
This loss further ensures that the reconstructed image 
maintains the same identity and facial pose as the original one. 
The final loss objective for the training phase I is formulated as 
\begin{equation}
    \mathcal{L}_\mathrm{P1} = \lambda \cdot \mathcal{L}_\mathrm{VQ} + \mathcal{L}_\mathrm{Sem},
\end{equation}
with $\lambda$ empirically set to 0.01 following~\cite{nitzan2020tog}.

\subsubsection{Training Phase II}
The second training phase involves a more elaborate strategy to guarantee 
the desired properties of anonymity, diversity, reversibility, and security, 
featured by the following loss objectives.

\paragraph{Anonymization loss} 
We first define anonymity loss to minimize the identity similarity 
between the anonymized and the original faces: 
\begin{equation}\label{eq:loss_anon}
    \mathcal{L}_\mathrm{Anon} = \mathbb{E}_{(\bm{X})} \theta^{+}\left(\bm{\Hat{X}}, \bm{X}\right),
\end{equation}
with $\theta^{+}$ defined as 
\begin{equation}
    \theta^{+}\left(\bm{X}, \bm{Y}\right) = \max \Big\{ 0, \theta\left(E_\mathrm{ID}\left(\bm{X}\right), E_\mathrm{ID}\left(\bm{Y}\right)\right) \Big\},
\end{equation}
where $\theta$ signifies cosine similarity between two embeddings. 

\paragraph{Diversity loss}
We then propose a diversity loss to boost the distinction on identity information between 
anonymized faces of multiple people using the same secret key ({\it inter-identity diversity}), 
and that of a single person using different keys ({\it inter-key diversity}):
\begin{equation}  
\mathcal{L}_{\text{Div}} = \mathbb{E}_{(\bm{\ddot{X}}_i\neq \bm{\ddot{X}}_j, s)} \theta^{+}(\bm{\ddot{X}}_i|s, \bm{\ddot{X}}_j|s) + \mathbb{E}_{(\bm{\ddot{X}}, s_k \neq s_l)} \theta^{+}(\bm{\ddot{X}}|s_k, \bm{\ddot{X}}|s_l)  
\end{equation} 
where $\bm{\ddot{X}}_i|s$ and $\bm{\ddot{X}}_j|s$ denote the anonymized faces of two people $i$ and $j$ using the same secret $s$, 
and $\bm{\ddot{X}}|s_k$ and $\bm{\ddot{X}}|s_l$ denote the anonymized faces of the same face using two different secrets $s_k$ and $s_l$. 

\paragraph{De-anonymization loss} 
For de-anonymization, we consider both scenarios involving matching and non-matching secrets. 
In the first scenario, a correctly recovered image $\bm{\check{X}}$ 
should closely mirror the original image in terms of facial identity. 
Conversely, any recovery image $\bm{\ddot{X}}$ generated using a non-matching 
secret should exhibit a significantly different identity from the original. 
The de-anonymization loss is therefore defined as
\begin{equation}\label{eq:loss_recid}
    \mathcal{L}_\mathrm{DeAnon} = \mathbb{E}_{(\bm{X})} \left[ \theta^{-}\left(\bm{\check{X}}, \bm{X} \right) + \theta^{+}\left(\bm{\ddot{X}}, \bm{X} \right) \right],
\end{equation}
where $\theta^{-}$ is defined as the inverse of cosine similarity:
\begin{equation}
    \theta^{-}\left(\bm{X}, \bm{Y} \right) = 1 - \theta\left(E_\mathrm{ID}\left(\bm{X}\right), E_\mathrm{ID}\left(\bm{Y}\right) \right).
\end{equation}

\paragraph{Multi-granularity image loss} 
Motivated by FIT~\cite{gu2020eccv}, we devise a multi-granularity image loss, 
applied on any generated image (denoted as $\bm{\bar{X}}$) regardless of its form 
(anonymized, recovered, or even falsely recovered), 
with the aim of maximizing their generation quality: 
\begin{equation}
    \mathcal{L}_\mathrm{Img} =\mathbb{E}_{(\bm{X})}  \Big[ \lambda_1\|\bm{X} - \bm{\bar{X}}\|_1 + \lambda_2 \mathrm{LPIPS}\left(\bm{X}, \bm{\bar{X}}\right) 
    + \lambda_3\| E_\mathrm{LM}\left(\bm{X}\right) - E_\mathrm{LM}\left(\bm{\bar{X}}\right)\|_1 
    + \lambda_4\| E_\mathrm{FP}\left(\bm{X}\right) - E_\mathrm{FP}\left(\bm{\bar{X}}\right)\|_2 \Big],
\end{equation}
where 
$\mathrm{LPIPS}$ denotes the Learned Perceptual Image Patch Similarity~\cite{richard2018lpips}, 
and $E_\mathrm{FP}$ represents a pretrained face parser from~\cite{lee2020maskgan}. 
Among all the loss weights $\bm{\lambda} = \{\lambda_1,\lambda_2,\lambda_3,\lambda_4\}$, 
$\lambda_1$ and $\lambda_2$ vary depending on the generated image type, 
whereas $\lambda_3$ and $\lambda_4$ are fixed:
\begin{equation}
    \bm{\lambda} = 
    \left\{ \begin{array}{lcl}
         10, \ 1, \ 0.1, \ 0.01 & \mbox{for} & \bm{\bar{X}} = \bm{\Check{X}}, \\ 
         0.01, \ 0.1, \ 0.1, \ 0.01 & \mbox{for} & \bm{\bar{X}} \in \{ \bm{\ddot{X}}, \bm{\Hat{X}} \}.
    \end{array}\right.
\end{equation}
The L1 and LPIPS metrics are assigned much lower weights for anonymized ($\bm{\ddot{X}}$) 
or falsely recovered ($\bm{\Hat{X}}$) images, as these images should primarily resemble 
the original image in terms of facial pose ($E_\mathrm{LM}$) 
and outline ($E_\mathrm{FP}$), rather than fine-grained details.

Finally, the loss objective for training phase II is defined as
\begin{equation}
    \mathcal{L}_\mathrm{P2} = \mathcal{L}_\mathrm{Anon} + \mathcal{L}_\mathrm{Div} + \mathcal{L}_\mathrm{DeAnon} + \mathcal{L}_\mathrm{Img}.
\end{equation}

\begin{figure}[h]
\centering
\includegraphics[width=0.9\columnwidth]{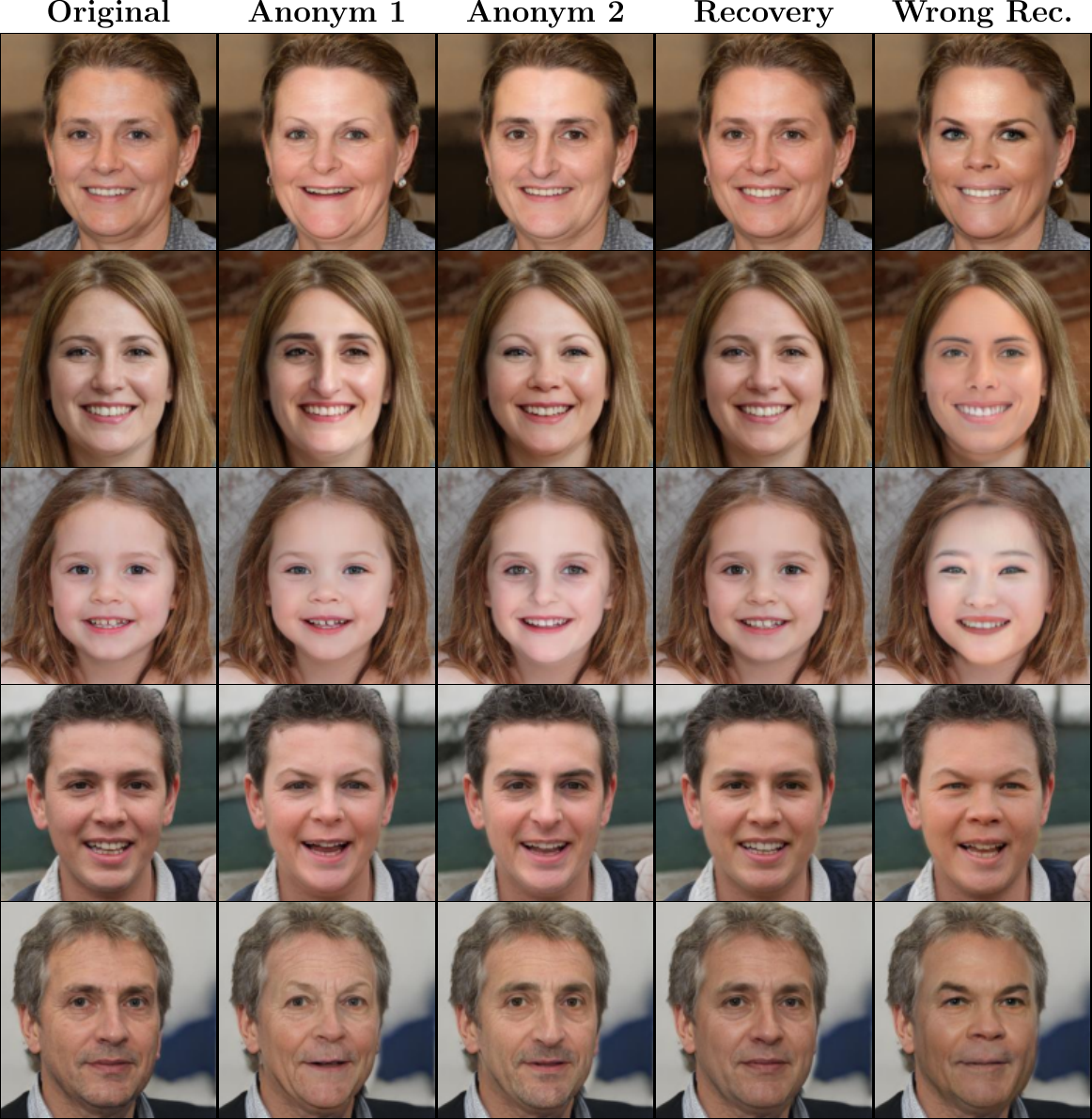}
\caption{Image samples of anonymization and de-anonymization results using different secret keys.}
\label{fig:overview_samples}
\end{figure}

\section{Experiments} \label{sec:experiments}

\subsection{Experimental Settings}
We primarily used the image dataset provided by~\cite{nitzan2020tog}, 
named as FFHQ*, which consists of 70,000 StyleGAN generated face images. 
simulating the Flickr-Faces-HQ Dataset (FFHQ)~\cite{karras2019style-ffhq}. 
We used 90\% of the dataset for training and the rest for testing.
In addition, we also included LFW~\cite{huang2007lfw}, FFHQ~\cite{karras2019style-ffhq} 
and CelebA-HQ~\cite{karras2017progressive-celebAHQ} in testing. 
All images in the experiments are with the resolution of 256$\times$256.

For the StyleGAN model, we used its second version, StyleGAN2, as proposed 
in~\cite{karras2020analyzing-stylegan2}. 
We obtained the backbone pretrained on FFHQ~\cite{karras2019style-ffhq} from the official 
open-source repository provided by the original authors~\cite{karras2020analyzing-stylegan2}. 
For model training, we used an Adam optimizer with $\beta_1=0$ and $\beta_2=0.999$ 
for both training phases, with a learning rate of $4\times10^{-4}$. 
The dual-phase training took approximately 150,000 iterations to converge, 
with a batch size of 4.
All experiments were conducted on a single NVIDIA RTX A6000 GPU. 
Several examples of anonymization and de-anonymization results 
utilizing different secrets are shown in Fig.~\ref{fig:overview_samples}.

\begin{figure}[ht]
     \centering
     \includegraphics[width=\columnwidth]{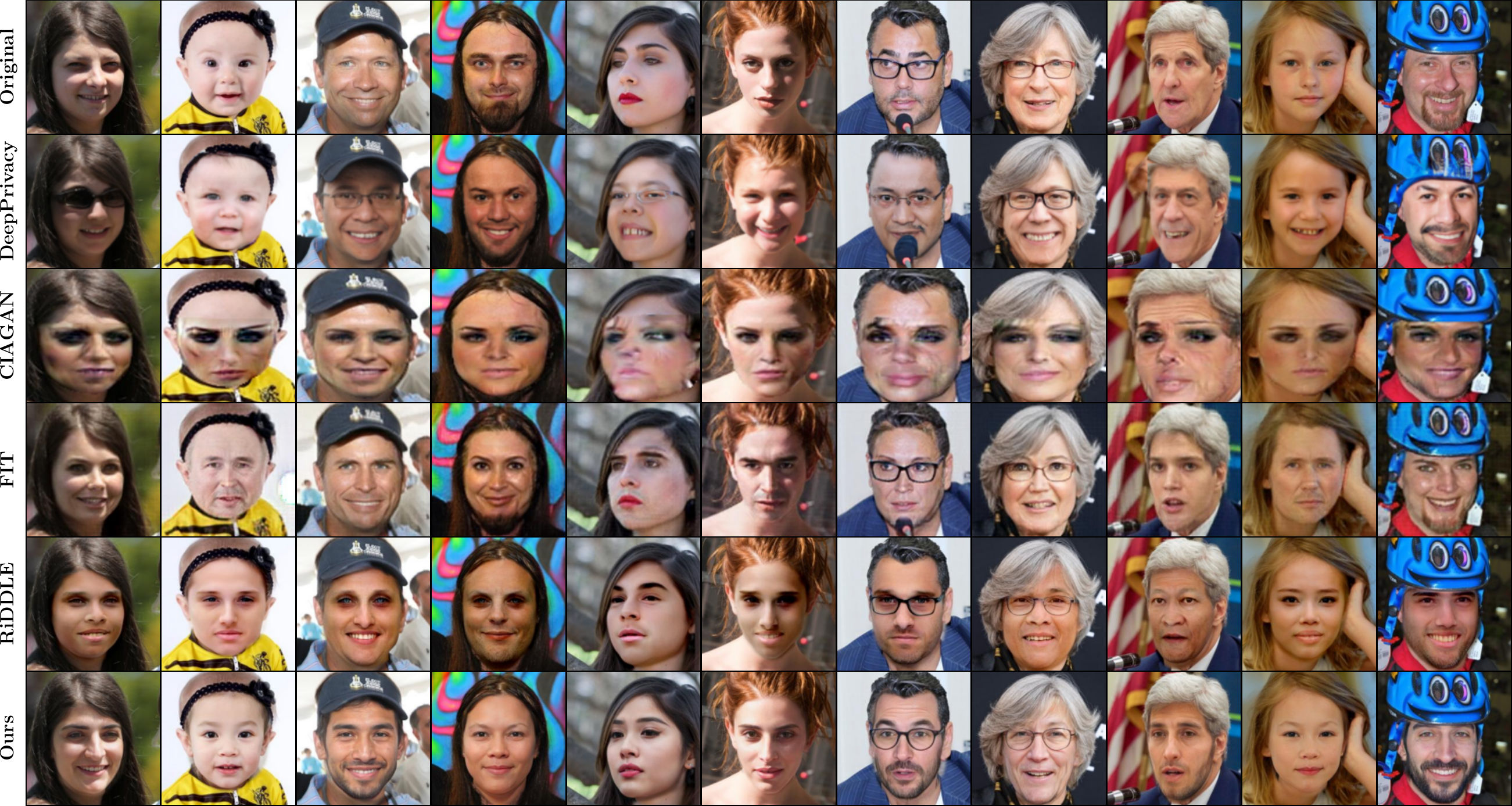}
     \caption{Qualitative comparison of the anonymization performance of various literature approaches. Original images are randomly selected from CelebA-HQ.}
     \label{fig:anonymized_images}
 \end{figure}

\subsection{Anonymization Performance}
We first analyze the anonymization performance of the proposed approach 
in comparison with existing methods including 
DeepPrivacy~\cite{hukkelas2019deepprivacy}, CIAGAN~\cite{maximov2020ciagan}, 
FIT~\cite{gu2020eccv}, and RiDDLE~\cite{li2023riddle}. 

\begin{table}[]
\centering
\tabcolsep=0.4em 
\caption{De-identification performance in terms of identity similarity (and distance) between the anonymized and original faces.}
\begin{tabular}{c|cccc|cccc}
\toprule
\multirow{3}{*}{Method} & \multicolumn{4}{c|}{CelebA-HQ~\cite{karras2017progressive-celebAHQ}}                             & \multicolumn{4}{c}{LFW~\cite{huang2007lfw}}                                   \\ \cmidrule(lr){2-5} \cmidrule(lr){6-9} 
                  & \multicolumn{2}{c}{ArcFace~\cite{deng2019arface}} & \multicolumn{2}{c|}{AdaFace~\cite{kim2022adaface}} & \multicolumn{2}{c}{ArcFace~\cite{deng2019arface}} & \multicolumn{2}{c}{AdaFace~\cite{kim2022adaface}} \\ \cmidrule(lr){2-3} \cmidrule(lr){4-5} \cmidrule(lr){6-7} \cmidrule(lr){8-9}
                  & Cos $\downarrow$        & L1 $\uparrow$          & Cos $\downarrow$        & L1 $\uparrow$          & Cos $\downarrow$        & L1 $\uparrow$          & Cos $\downarrow$        & L1 $\uparrow$          \\ \midrule 
CIAGAN        & 0.587 & 0.016    &0.234       & 0.022          & 0.341        & 0.020    & 0.154                         & 0.023               \\ 
DeepPrivacy     & 0.531        & 0.017   & 0.172        & 0.023          & 0.555     & 0.017         & 0.373                       & 0.019            \\ 
FIT     & 0.125       & 0.033    & 0.100        & 0.023          & 0.272       & 0.032     & 0.159                    & 0.023           \\ 
RiDDLE       & 0.099             & 0.024   & 0.076       & 0.024          & 0.148        & 0.023   & 0.101                 & 0.024              \\ \midrule
Ours   & \textbf{0.061}       & \textbf{0.048} & \textbf{0.048}   & \textbf{0.024} & \textbf{0.020}  & \textbf{0.049}& \textbf{0.015}        & \textbf{0.025}         \\ 
\bottomrule
\end{tabular}
\label{tab:cos}
\end{table}

Qualitative anonymization samples from different methods are 
visualized in Fig.~\ref{fig:anonymized_images}. 
The proposed approach generates the most natural anonymization appearance, 
with optimal visual quality and the highest coherence with the original image, 
while presenting significant differences in visual identity. 
In contrast, the literature approaches exhibit more or less 
artifacts related to illumination, facial boundaries, or skin color. 

Quantitatively, we evaluate the face de-identification performance by inspecting 
the similarity in identity embeddings between anonymized and original faces, 
using multiple identity extractors (ArcFace~\cite{deng2019arface} 
AdaFace~\cite{kim2022adaface}) and distance metrics (cosine similarity and L1 distance). 
Lower similarity or higher distance implies a stronger capability to prevent 
anonymized faces from being correctly recognized by machines. 
As shown in the de-identification results in Table~\ref{tab:cos}, 
the proposed approach offers the strongest de-identification performance 
compared to existing methods on both the CelebA-HQ and LFW datasets, 
as indicated by all tested face recognizers and distance metrics. 

To verify that anonymized faces remain applicable for downstream tasks, 
we use multiple indices to evaluate their utility. 
These include the face detection success rate (proportion of correctly detected faces), 
the accuracy of the detected facial bounding box location (L1 distance of the four coordinates), 
the precision of detected facial landmarks (L1 distance of the 68 key points), 
and the Fréchet Inception Distance (FID)~\cite{heusel2017fid}, 
which measures the overall distribution similarity between the original and anonymized images.
The results, compared to existing methods, are shown in Table~\ref{tab:utility}. 
The anonymized images produced by the proposed approach achieve 
the highest scores across all the utility factors mentioned above.
These results align with the qualitative comparisons shown in Fig.~\ref{fig:anonymized_images}, 
demonstrating the optimal visual quality and facial naturalness of the synthesized images by the proposed approach.

\begin{table}
\tabcolsep=0.1cm 
\centering
\caption{Quantitative results indicating the utility of anonymized images for different approaches, 
measured on CelebA-HQ~\cite{karras2017progressive-celebAHQ}.}
\begin{tabular}{cccccccc} 
\toprule
\multirow{2}{*}{Method} & \multicolumn{2}{c}{Face Detection $\uparrow$} & \multicolumn{2}{c}{Detection Box $\downarrow$} & \multicolumn{2}{c}{Landmark $\downarrow$} & \multirow{2}{*}{FID $\downarrow$}    \\ 
\cmidrule(r){2-3} \cmidrule(r){4-5} \cmidrule(r){6-7}
            & MtCNN          & Dlib              & MtCNN  & Dlib                             & MtCNN          & Dlib                &         \\ 
\midrule
DeepPrivacy & 1.000 & 0.980         & 4.654  & 2.685           & 3.280  & 2.896               & 23.713  \\
CIAGAN      & 0.992 & 0.937         & 20.387 & 15.476          & 8.042  & 8.930               & 32.611  \\ 
FIT         & 1.000 & 0.984         & 7.879  & 4.218           & 3.572  & 4.047               & 30.331  \\ 
RiDDLE      & 1.000 & 0.991         & 3.824  & 1.700           & 1.674  & 1.512               & 15.389  \\ 
\midrule
\bf{Ours}        & \textbf{1.000} & \textbf{0.995}             & \textbf{0.998}  & \textbf{0.850}                            & \textbf{1.219} & \textbf{0.986}      & \textbf{14.760}         \\
\bottomrule
\end{tabular}
\label{tab:utility}
\end{table}

\begin{figure*}[!t]
     \centering
     \includegraphics[width=\textwidth]{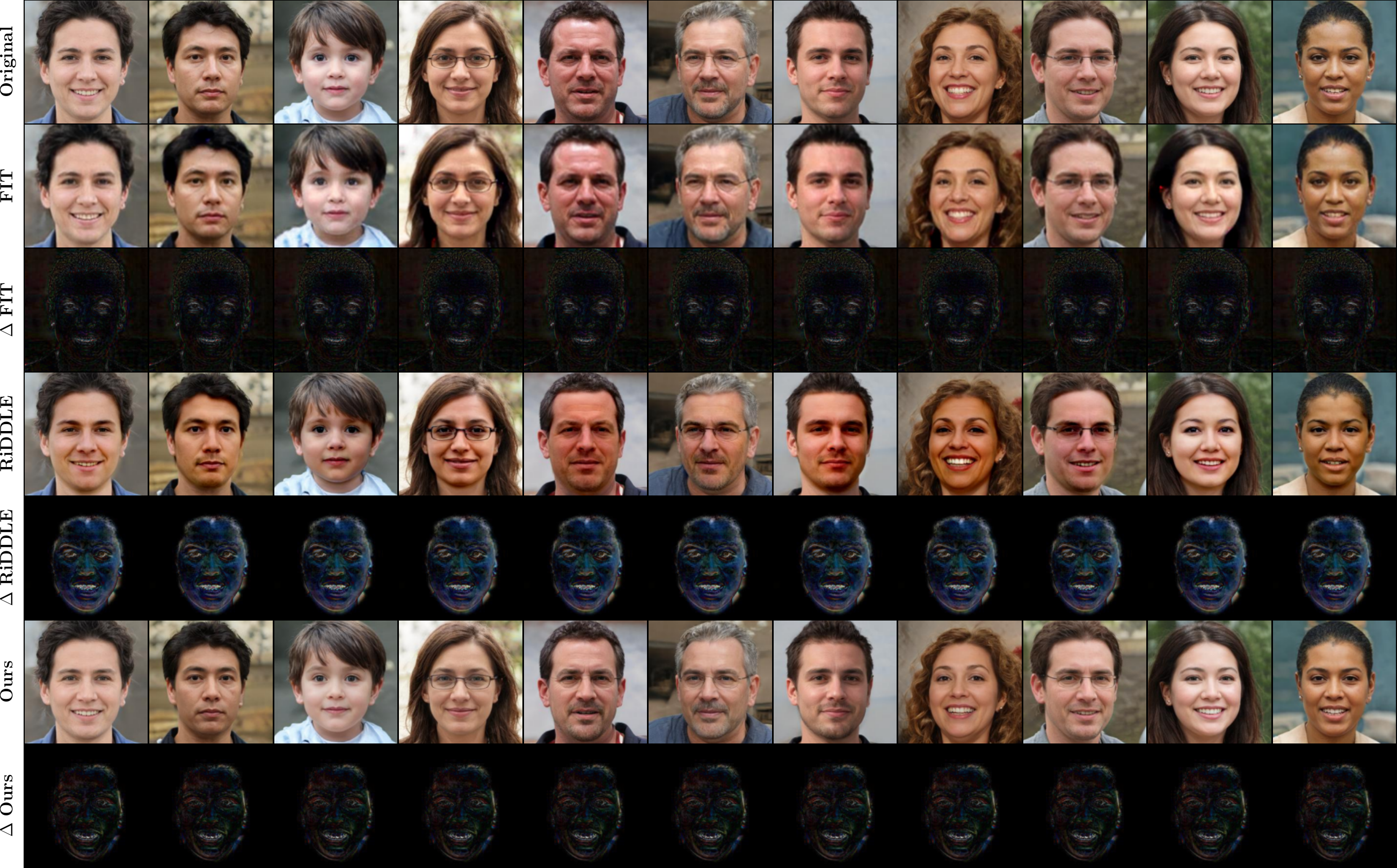}
     \caption{Qualitative comparison of the de-anonymization image quality of literature approaches that support reversibility. Original images are randomly selected from the StyleGAN synthesized dataset FFHQ*~\cite{nitzan2020tog}. $\Delta$ indicates the $2\times$ magnified absolute difference between the recovery image and the original image.}
     \label{fig:plot_compare_recovery_images} 
 \end{figure*}

\begin{table}[t]
\caption{Comparison of de-anonymization quality among literature approaches, measured using multiple metrics.}
\centering
\tabcolsep=0.5em
\begin{tabular}{ccccccc} 
\toprule
       \multirow{2}{*}{} & \multirow{2}{*}{MSE $\downarrow$} & \multirow{2}{*}{PSNR $\uparrow$} & \multirow{2}{*}{SSIM $\uparrow$} & \multirow{2}{*}{LPIPS $\downarrow$} & \multicolumn{2}{c}{Cosine similarity $\uparrow$} \\ \cmidrule{6-7}
                         &                    &                    &                    &                    & ArcFace        & AdaFace       \\ 
\midrule
FIT    &\underline{3.9e-4} 	        &\underline{28.08}       &0.896  	&0.151 & \textbf{0.917} & \textbf{0.882} 	\\
RiDDLE &7.7e-4 	        &25.39       &\underline{0.928}  	&\underline{0.073} & 0.804 & 0.705 	\\
\midrule
\bf{Ours}   &\textbf{3.2e-4}	&\textbf{35.20}      &\textbf{0.945} 	&\textbf{0.041}	& \underline{0.852}	& \underline{0.786}	\\
\bottomrule
\label{tab:recovery_quality}
\end{tabular}
\end{table}

\subsection{De-anonymization Performance}
We then evaluate the de-anonymization performance of the proposed method 
by inspecting the similarity between correctly de-anonymized images and 
the corresponding original images, both qualitatively and quantitatively. 
First, exemplary recovery images from different reversible approaches, 
including the differential image between the recovered and original ones, 
are shown in Fig.~\ref{fig:plot_compare_recovery_images}. 
The proposed approach shows a significant improvement 
compared to the recent RiDDLE~\cite{li2023riddle}, 
offering more accurate reconstruction of facial details and image color. 
Although our results demonstrate comparable quality to FIT~\cite{gu2020eccv}, 
the recovery images from the proposed method exhibit higher resolution 
and finer details, due to the application of StyleGAN. 
Quantitatively, we computed the similarity between the de-anonymized 
and original images using multiple indexes, including image quality metrics such as 
MSE, PSNR, SSIM~\cite{wang2004ssim}, and LPIPS~\cite{richard2018lpips}, 
and cosine similarity based on identity embedding. 
The results are summarized in Table~\ref{tab:recovery_quality}. 
It is evident that the de-anonymization quality of the proposed approach surpasses 
that of the other methods across all image quality metrics, consistent with the 
subjective observations shown in Fig.~\ref{fig:plot_compare_recovery_images}. 
The cosine similarity in facial identity between de-anonymized and original images 
remains high for both face recognizers, although our results are slightly worse than FIT. 
This is due to the slight semantic changes introduced by StyleGAN.

\def\weight{0.24}
\begin{figure*}[t]
\centering
  \subfloat
  {\includegraphics[height=\weight\textwidth]{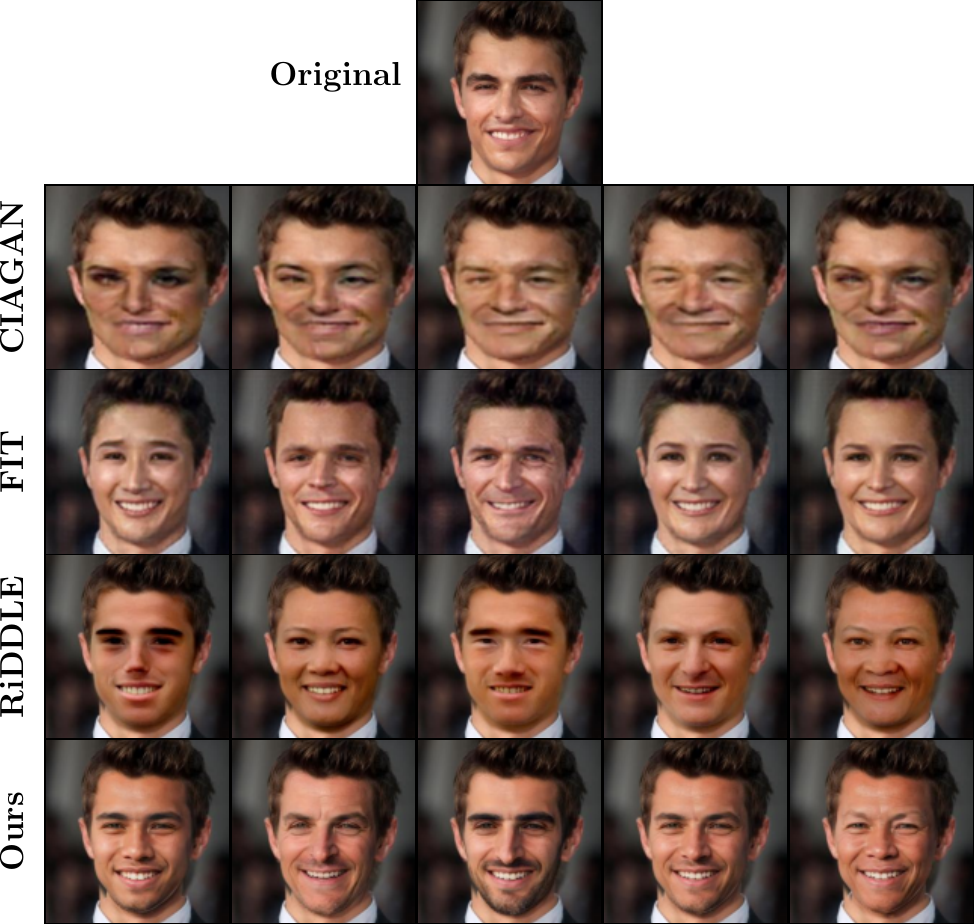}}
  \hfil
  \subfloat
  {\includegraphics[height=\weight\textwidth]{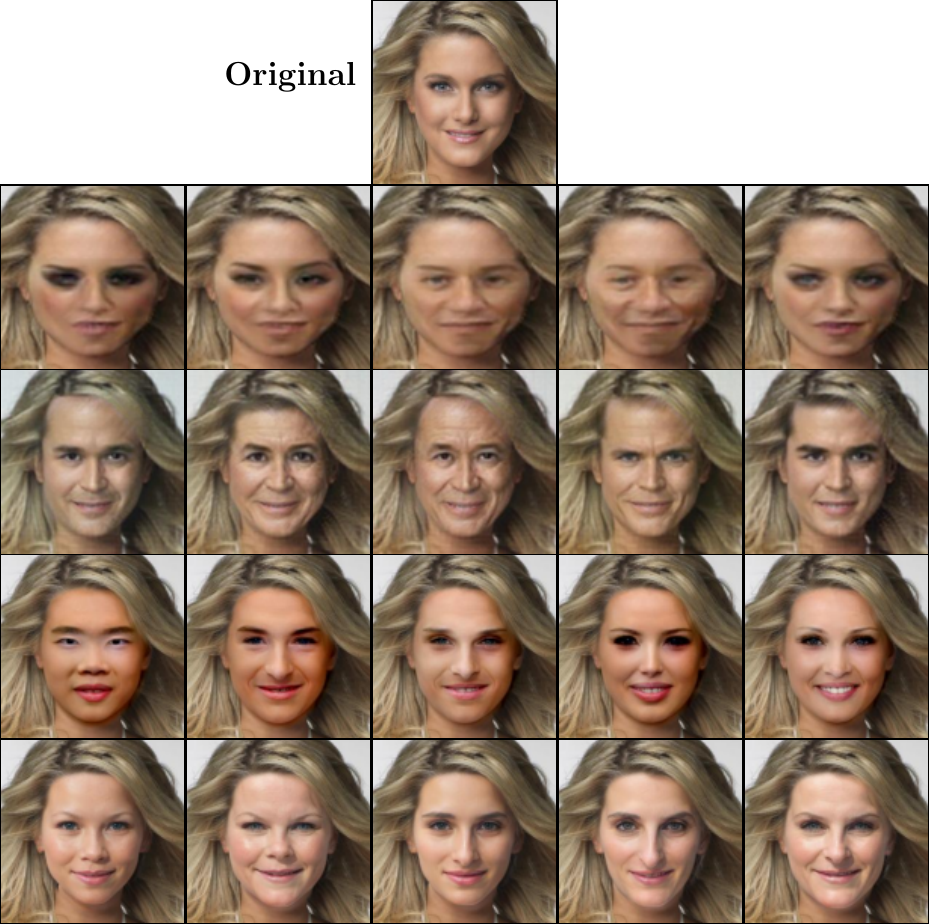}}
  \hfil
  \subfloat
  {\includegraphics[height=\weight\textwidth]{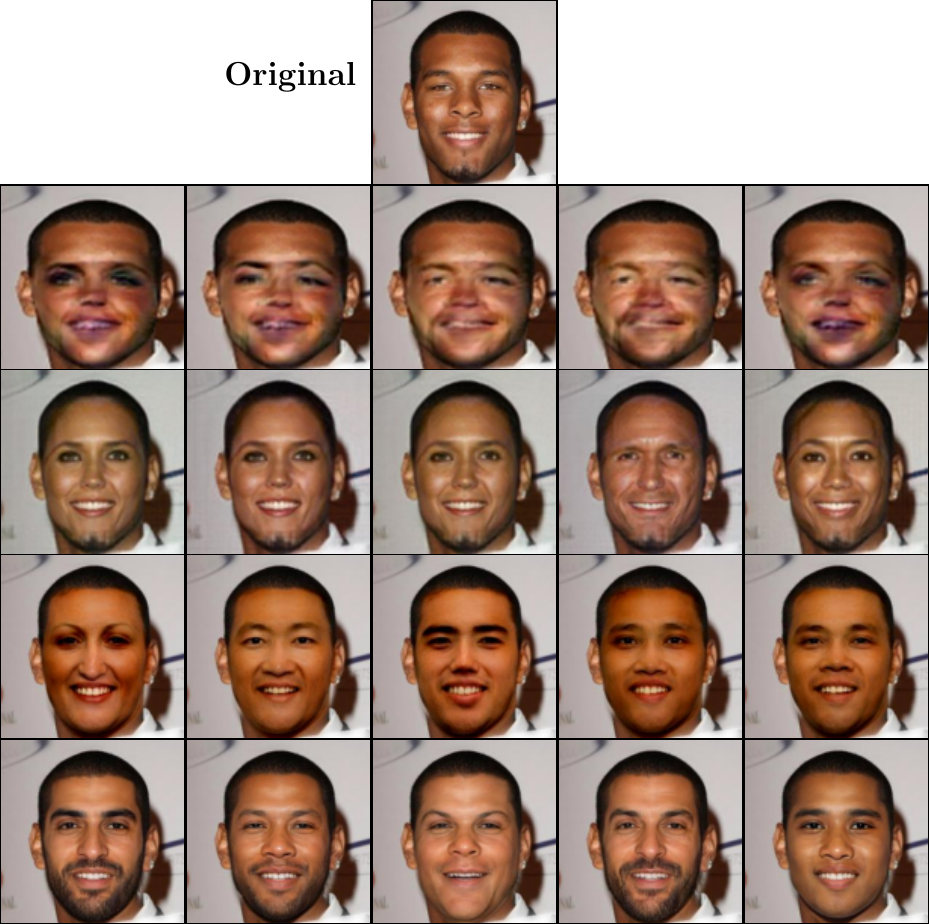}}
  \hfil 
  \subfloat
  {\includegraphics[height=\weight\textwidth]{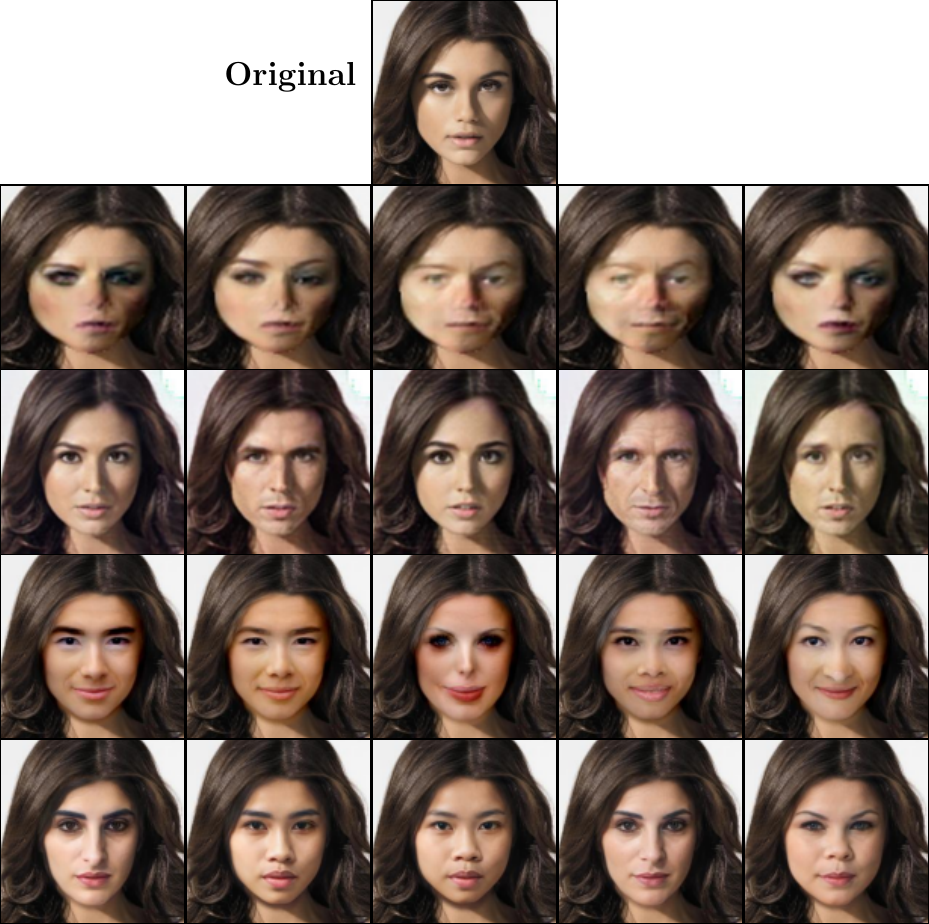}}
  \caption{Qualitative evaluation of anonymization diversity: multiple anonymized samples (across columns) corresponding to different secret keys for various approaches (across rows) are visualized. The proposed approach demonstrates the highest level of both diversity and naturalness.}
\label{fig:diversity_samples}
\end{figure*}

\subsection{Security and Diversity Analysis}
To ensure security, anonymized faces corresponding to different secret keys 
are expected to represent distinct identities as much as possible. 
Similarly, de-anonymized faces using different keys should also exhibit 
significant differences in identity. This property, often referred to as diversity, 
is highly relevant to the security of an anonymization method.
To demonstrate the diversity of the proposed approach, 
we showcase quantitative and qualitative results compared with 
literature methods in Table~\ref{tab:diversity} and 
Fig.~\ref{fig:diversity_samples} respectively. 
For each test image, we generated multiple anonymized and de-anonymized 
examples using several random secret keys. 
We calculated the average cosine similarity in facial identities between the anonymized and original faces, 
along with the corresponding Known Factor Feature Angle (KFFA)~\cite{kffa} score. 
The KFFA score was computed based on a pretrained ArcFace~\cite{deng2019arface} model 
and is typically used as an indicator of anonymization diversity.
It is evident that the proposed approach generates significantly more diverse results 
compared to CIAGAN~\cite{maximov2020ciagan} and FIT~\cite{gu2020eccv}. 
The recent RiDDLE~\cite{li2023riddle} shows comparable diversity to ours, 
but with noticeably lower facial naturalness, as observed in Fig.~\ref{fig:diversity_samples}.

\begin{table}
\centering
\caption{Comparison of anonymization diversity among literature approaches based on  
average cosine similarity and KFFA~\cite{kffa}.}
\begin{tabular}{cccccc}
\toprule
Method                  & Cosine similarity $\downarrow$ & KFFA $\uparrow$ \\ 
\midrule
CIAGAN                  & 0.254     & 21.99     \\ 
FIT                     & 0.192     & 54.18     \\
RiDDLE                  & 0.138     & 71.64     \\
\midrule
\bf{Ours (anonymized)}       & \textbf{0.062}     & \textbf{74.72}       \\
\bf{Ours (de-anonymized)}    & 0.172       & 70.80       \\
\bottomrule
\end{tabular}
\label{tab:diversity}
\end{table}

\begin{figure}[t]
\centering
  \includegraphics[width=0.9\columnwidth]{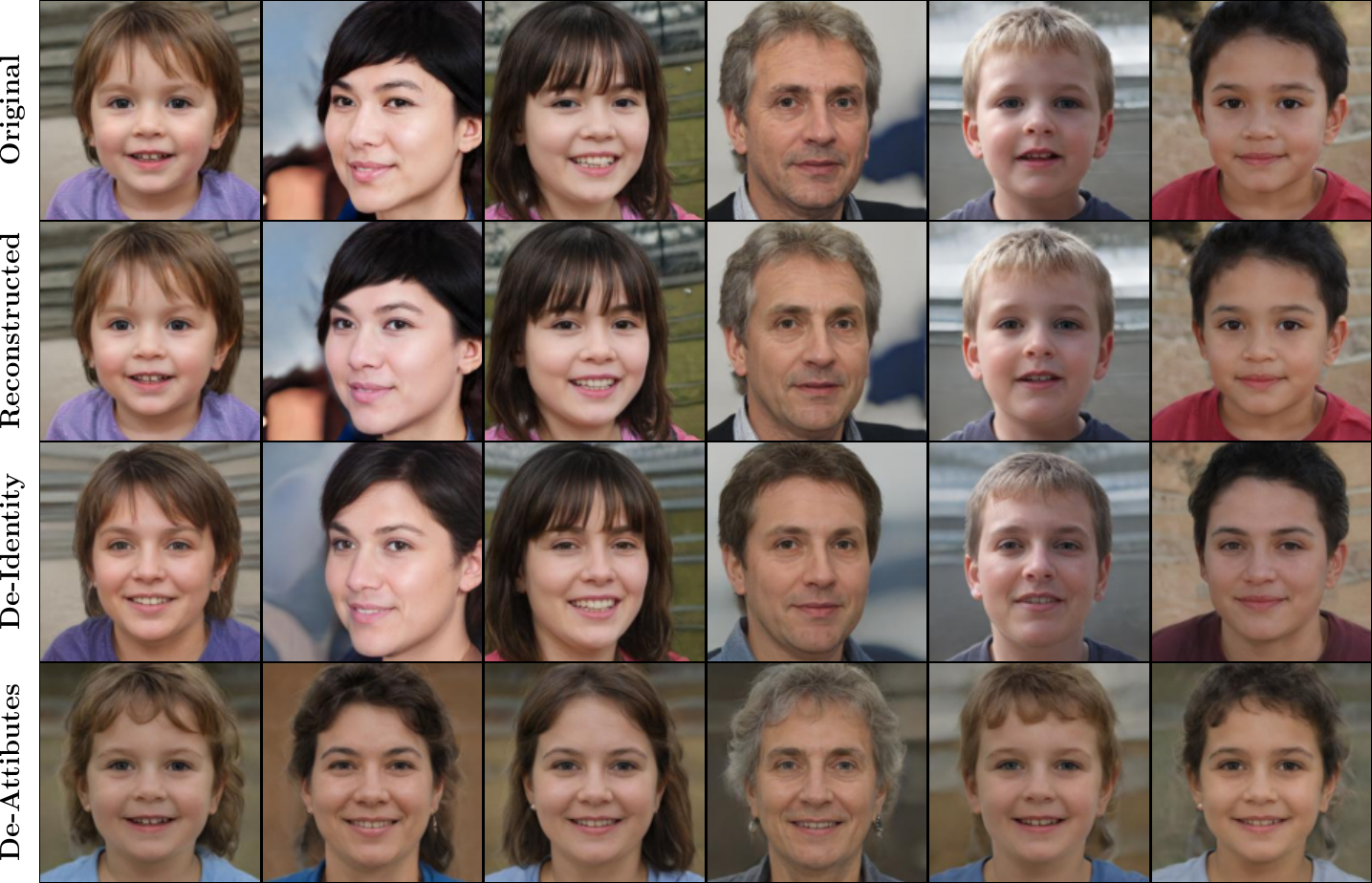}
  \caption{Qualitative analysis of the disentanglement and reconstruction architecture. 
  The four rows from top to bottom denote: the original images, 
  the reconstructed images from both disentangled features, 
  the reconstructed images from only attribute features ({\it De-Identity}), 
  and the reconstructed images from only identity features ({\it De-Attributes}), respectively.}
\label{fig:plot_disentangled_effect}
\end{figure}

\subsection{Effectiveness of Disentanglement}
To showcase the effect of identity disentanglement, 
we analyze the contribution of each decoupled representation. 
To this end, we extracted the disentangled representations and conducted 
image reconstruction without SIF, using only one of two representations. 
We set all values of the excluded identity or attributes representation to zero, 
denoted as {\it De-Identity} and {\it De-Attributes} respectively. 
The reconstruction results in various scenarios 
are visualized in Fig.~\ref{fig:plot_disentangled_effect}.  

In the {\it De-Identity} scenario (the 3rd row of Fig.~\ref{fig:plot_disentangled_effect}), 
where the identity information is removed, reconstructed images exhibit 
slight differences in identity compared to the original, 
while preserving most non-identifying attributes such as facial pose and background. 
Interestingly, the visual identities represented by multiple {\it De-Identity} images 
tend to be homogeneous, although they originally had different identities, ages, and genders. 
In the {\it De-Attributes} scenario (the 4th row of Fig.~\ref{fig:plot_disentangled_effect}), 
where the attributes information is removed, 
all reconstructed images exhibit almost the same frontal pose and background,
with similar haircuts regardless of gender. 
While, the visual identity represented by those {\it De-Attributes} 
images is still close to their original. 
This clearly demonstrates the capability of the decoupled attributes representation 
in representing the non-identifying information. 
When performing image reconstruction by combining both representations 
(the 2nd row of Fig.~\ref{fig:plot_disentangled_effect}), 
we obtain synthesized faces that closely approximate the original ones. 
This highlights the necessity of incorporating both disentangled features in image reconstruction. 

Given that the disentanglement architecture effectively decouples identity and attribute information, 
combined with the inherent invertibility of the flow-based model for identity transformation, 
we assert that the proposed approach offers enhanced interpretability compared to most literature methods.

\def\weight{0.249}
\begin{figure*}[t]
\centering
  \subfloat
  {\includegraphics[width=\weight\textwidth]{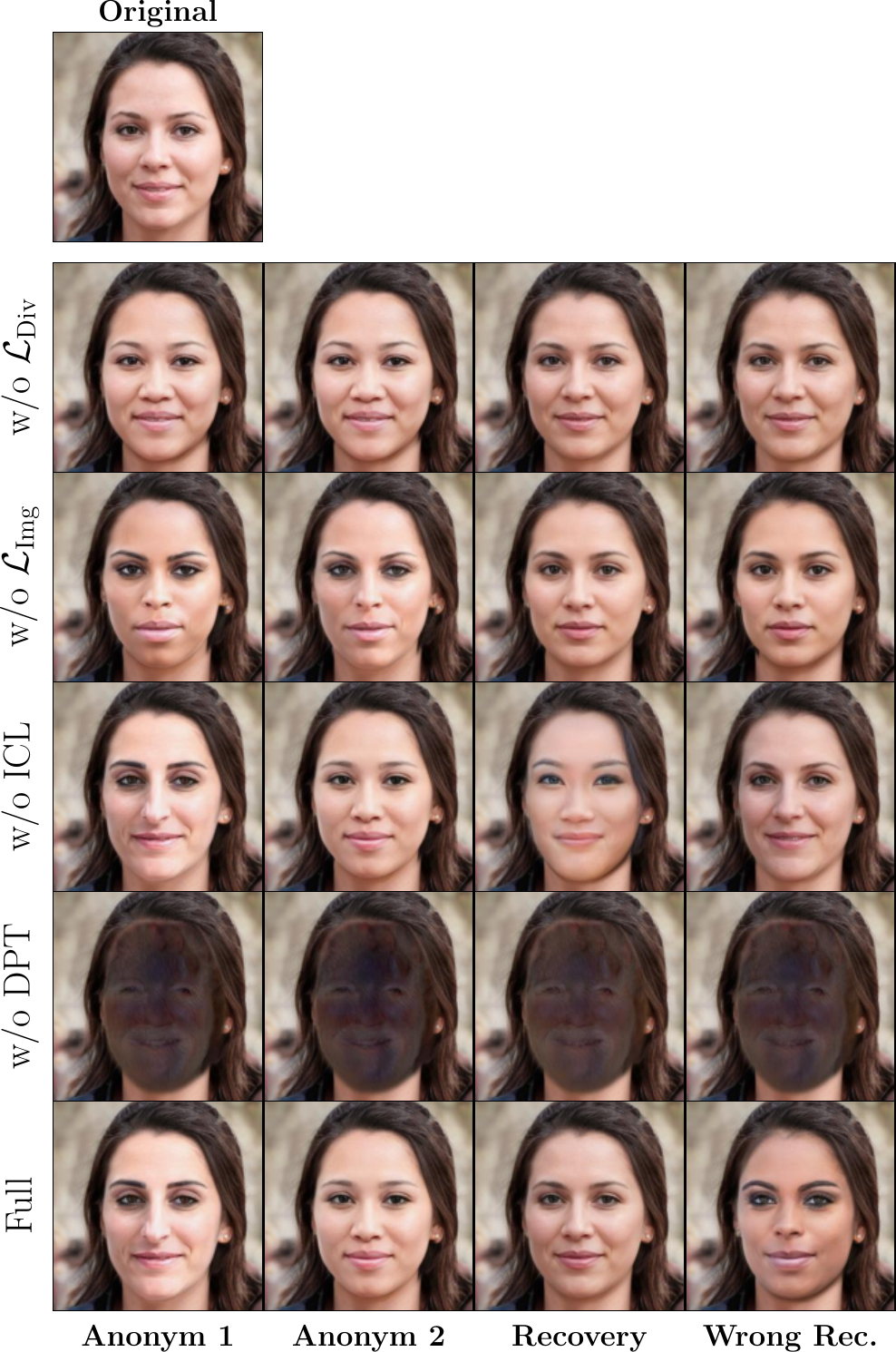}}
  \hfill
  \subfloat
  {\includegraphics[width=\weight\textwidth]{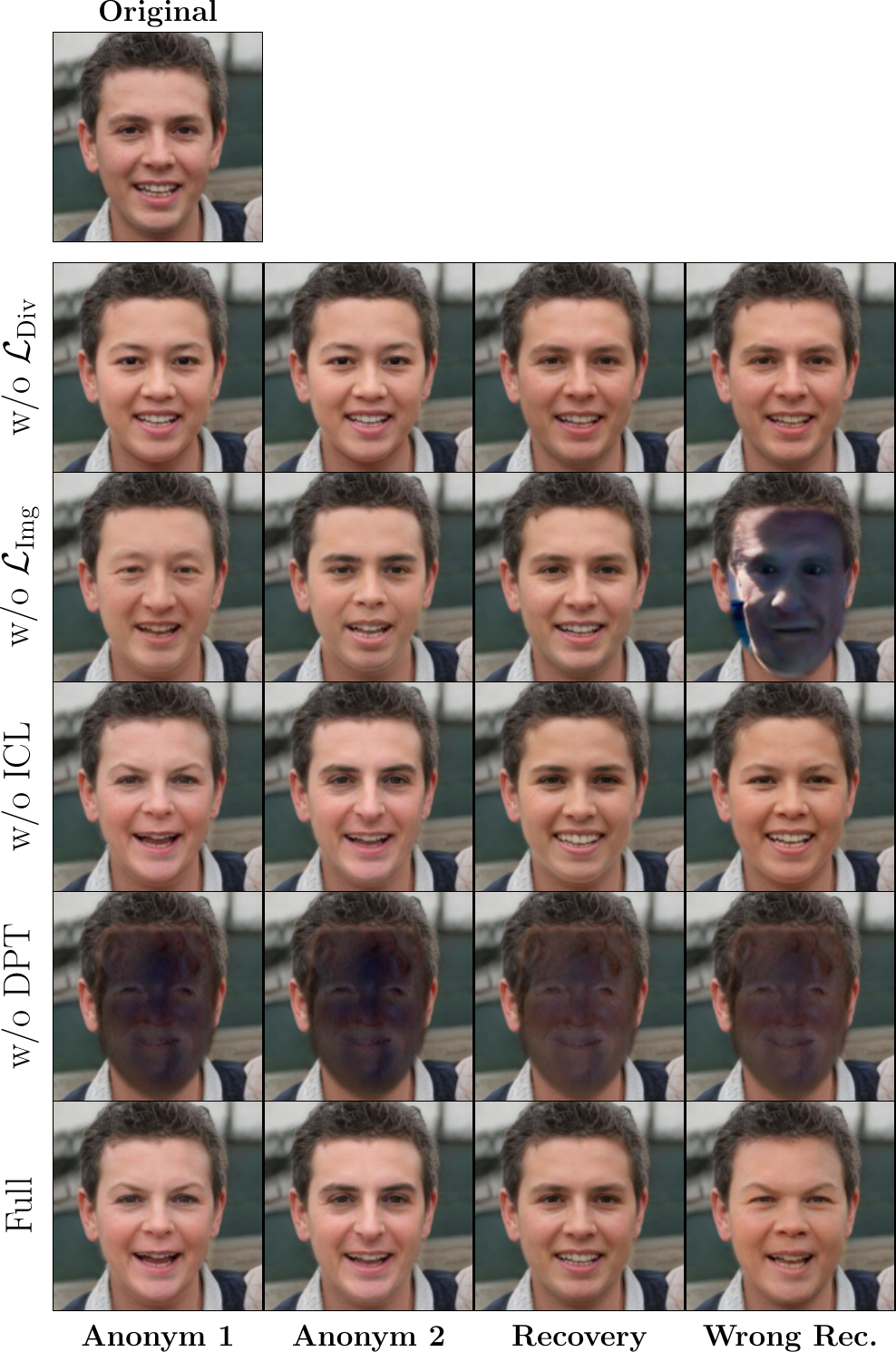}}
  \hfill
  \subfloat
  {\includegraphics[width=\weight\textwidth]{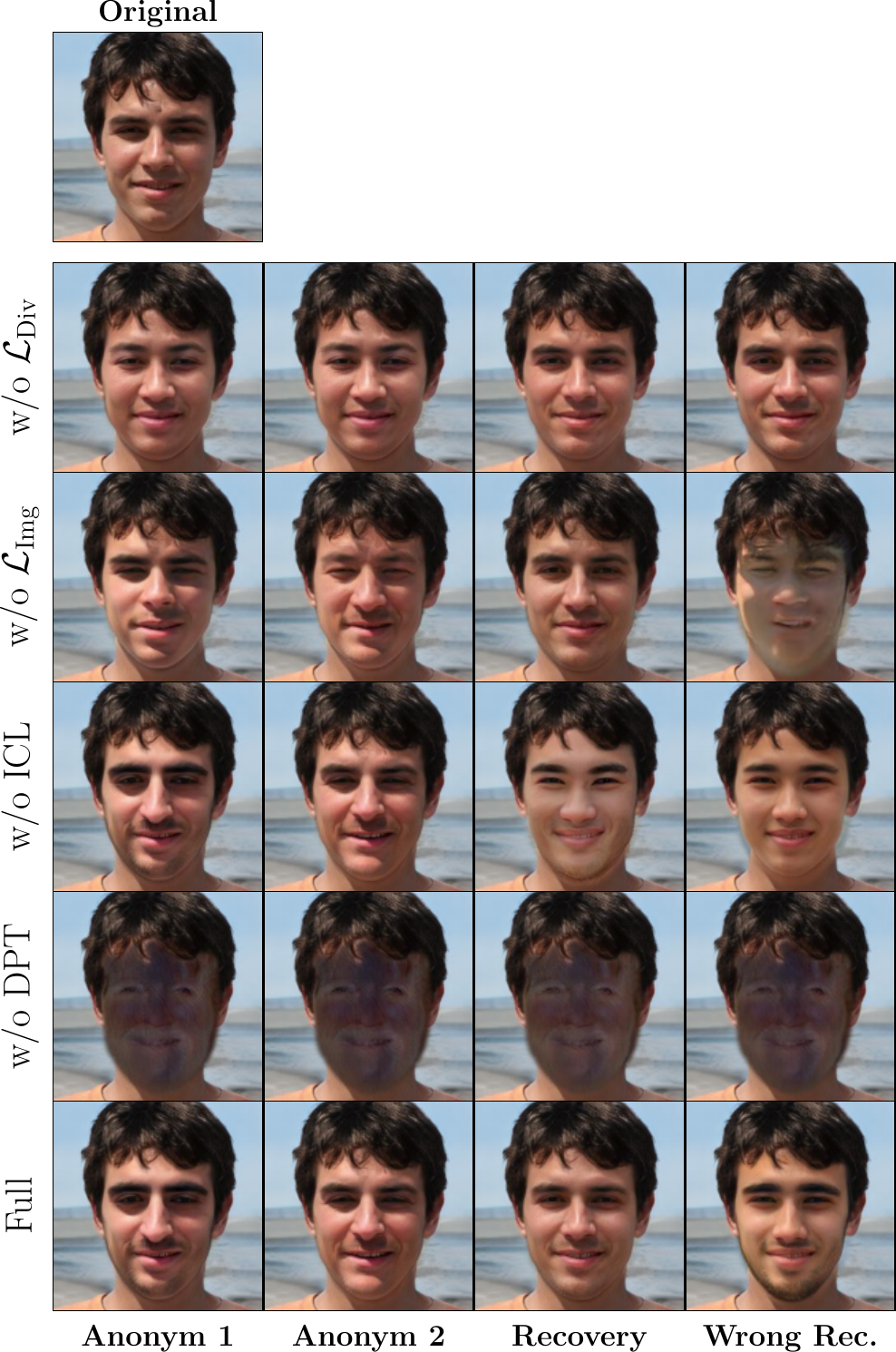}}
  \hfill
  \subfloat
  {\includegraphics[width=\weight\textwidth]{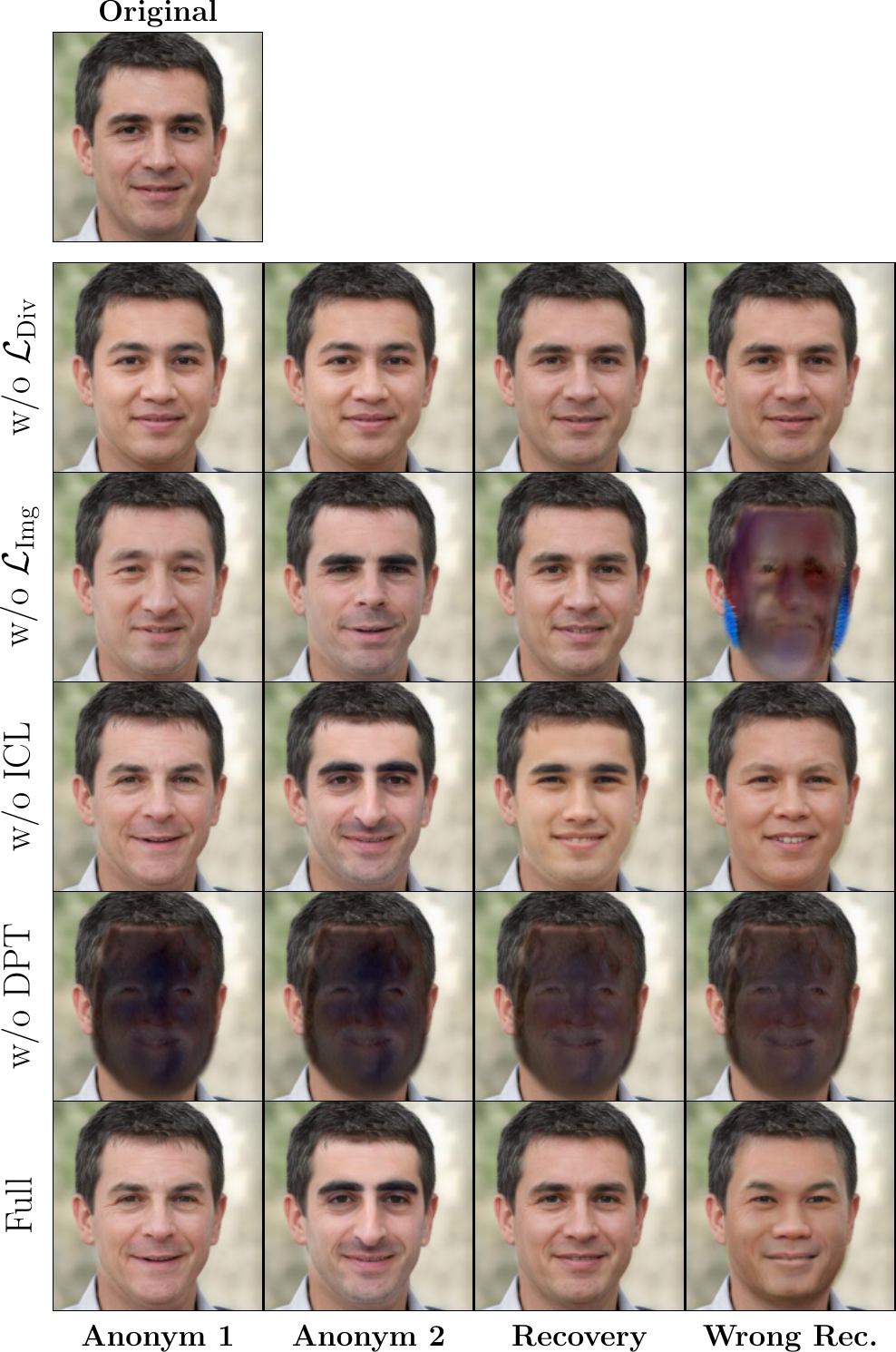}}
  \caption{Samples of anonymization and de-anonymization results obtained under different conditions of the ablation study.}
\label{fig:ablation_study}
\end{figure*}

\subsection{Ablation Study}
Last but not least, we verify the effectiveness of the key components 
with the proposed framework through a series of ablation experiments. 
Specifically, we examine the contribution of the proposed diversity loss 
$\mathcal{L}_\mathrm{Div}$, the multi-granularity image loss $\mathcal{L}_\mathrm{Img}$, 
and the introduced identity compensation layer (ICL), 
by rebuilding the proposed model when excluding each of those components. 
In particular, we also verified the effectiveness of the proposed dual-phase training (DPT) 
strategy by evaluating the performance of the model trained with all loss functions together 
within a single phase (w/o DPT). The anonymization and de-anonymization performance 
corresponding to these cases are reported in Table~\ref{table:ablation_study}. 
Meanwhile, we showcase anonymized and de-anonymized image samples qualitatively 
corresponding to the aforementioned cases in Fig.~\ref{fig:ablation_study} Qualitatively.


Without the diversity loss~(w/o~$\mathcal{L}_\mathrm{Div}$), 
the KFFA score of anonymized images is significantly reduced, 
highlighting its effectiveness in enabling diverse anonymizations. 
When removing the image quality loss~(w/o~$\mathcal{L}_\mathrm{Img}$), 
the overall quality of synthesized images is impacted, 
as reflected by the increased FID score and the degraded visual quality 
of recovered images (especially those false recovery cases) as shown in Fig.~\ref{fig:ablation_study}. 
Without the introduced identity compensation layer (w/o ICL) for compensating the identity information, 
the de-anonymization quality is greatly affected. 
The recovery images exhibit much lower cosine similarity compared to the original ones. 
Finally, if we include all loss functions within only a single training pipeline 
(w/o DPT), the training could not properly converge, resulting in highly contaminated synthesized faces.
This is illustrated in Fig.~\ref{fig:ablation_study} and 
reflected by the high FID scores shown in Table~\ref{table:ablation_study}.
Note that in this case, the average cosine similarity of anonymized faces is extremely low, 
simply because the generated faces are abnormal.

\begin{table}[t]
\caption{Results of the ablation study.}
\centering
\tabcolsep=0.13cm 
\begin{tabular}{lcccccc}
\toprule
\multirow{2}{*}{}       & \multicolumn{3}{c}{Anonymization} & \multicolumn{3}{c}{De-anonymization}  \\ 
\cmidrule(lr){2-4} \cmidrule(lr){5-7}
                        & Cosine $\downarrow$ & FID $\downarrow$  & KFFA $\uparrow$    & SSIM $\uparrow$  & LPIPS $\downarrow$  &Cosine $\uparrow$               \\ 
\midrule
w/o $\mathcal{L}_\mathrm{Div}$  & 0.195         & 14.940   & 23.069 & 0.945 & 0.040    & 0.954             \\ 
w/o $\mathcal{L}_\mathrm{Img}$  & 0.181        & 20.524  & 72.618  & 0.943 & 0.042    & 0.951       \\ 
w/o ICL                         & 0.208         & 14.760   & 69.957 & 0.916 & 0.069    & 0.771       \\ 
w/o DPT                         & \textbf{0.010}         & 195.476  & 19.897  & 0.837 & 0.233     & 0.010          \\ 
\midrule
FULL                    & 0.062        & \textbf{14.760}  & \textbf{73.165}  & \textbf{0.945} & \textbf{0.039}     & \textbf{0.954}       \\
\bottomrule
\end{tabular}
\label{table:ablation_study}
\end{table}

\section{Conclusion} \label{sec:conclusion}
In this paper, we propose a novel approach to invertible face anonymization 
by exploring a disentangled paradigm that decouples sensitive 
identity information from the image and transforms the 
decoupled identity in a reversible manner. 
To ensure secure transformation and reversibility, 
we devise a secret key conditioned flow-based model 
that enables secure anonymization and de-anonymization, 
allowing correct de-anonymization results only with the matching secret key. 
We introduce a dedicated dual-phase training strategy to progressively 
optimize the disentanglement and transformation processes, 
maximizing the desired functionalities. 
Qualitative and quantitative experiments conducted on 
multiple face image datasets demonstrate the effectiveness 
of the proposed approach in achieving improved anonymity, 
reversibility, security, diversity, and model interpretability, 
compared to most literature approaches.


Nevertheless, several limitations still remain. 
The disentanglement performance can be further improved, 
as the disentangled attribute representation still retains 
a certain amount of identity information, as observed in 
Fig.~\ref{fig:plot_disentangled_effect}. 
Additionally, similar to most StyleGAN-based anonymization methods, 
such as RiDDLE~\cite{li2023riddle}, it is challenging for the recovery image 
to preserve exactly the same semantic information as the original, 
due to the nature of StyleGAN. 
Addressing these issues will be the focus of our future work. 


\bibliographystyle{elsarticle-num} 
\bibliography{references.bib}

\end{document}